# Generative Memory for Lifelong Reinforcement Learning

Aswin Raghavan, Jesse Hostetler, Sek Chai[†]
SRI International, Princeton, NJ, USA

## ABSTRACT

Our research is focused on understanding and applying biological memory transfers to new AI systems that can fundamentally improve their performance, throughout their fielded lifetime experience. We leverage current understanding of biological memory transfer to arrive at AI algorithms for memory consolidation and replay. In this paper, we propose the use of generative memory that can be recalled in batch samples to train a multi-task agent in a pseudo-rehearsal manner. We show results motivating the need for task-agnostic separation of latent space for the generative memory to address issues of catastrophic forgetting in lifelong learning.

## KEYWORDS

Lifelong learning, generative auto-encoder, memory consolidation, pseudo-rehearsal, replay, reinforcement learning.

## 1   Introduction

While AI systems have become core to many cognitive applications today, they do not handle new scenarios that they are not trained on, and they can repeatedly make the same mistakes. Even with retraining, today's systems are prone to "catastrophic forgetting" when a new item disrupts previously learned knowledge. Our goal is to address these limitations by enabling AI system that can continually learn. Such a capability requires the AI system to know what to learn and when, and to utilize a memory subsystem to support unsupervised on-line learning.

Learning multiple tasks in sequence remains a challenge because standard AI system forget information related to previously learned tasks. While there are numerous methods proposed for alleviating catastrophic forgetting [1-4], most methods have not addressed issues for scalability. For example, progressive and ensemble neural networks [5,6] adds incremental layers to leverage previously learned features. However, such approaches grow the network parameters substantially to accommodate the new information. A network pruning step can be included, but such a process would be subject to forgetting.

We are developing a lifelong reinforcement learning (RL) approach that uses a memory subsystem to generate training samples for unsupervised learning. As shown in Figure 1, the replay memory is integrated into the RL framework, and can provide batch samples for the learning agent. Using Deep-Q learning algorithm, for example, the agent can recall a set of replay samples and retrain based on a new scenario.

Current RL approaches requires a large FIFO queue to store sufficient training samples. Furthermore, for replay-based training, there is additional requirement to know a-priori task labels to recall the proper samples. As such, current replay approach for RL are not scalable, and limiting, due to the requirement of large training corpus for lifelong learning.

We propose the use of a generative memory component [7-9] for the replay memory to support scalability for lifelong learning. We show an approach whereby the embedding of tasks and instances can be optimally separately in latent space such that task labels are no longer necessary for recall. We provide results for an example auto-encoder style method for the generative memory.

## 2   Background and Biological Inspiration

Biological memory transfer is a complex sequence of dynamic processes, with local and global synchronization patterns. These processes support memories with flexibility in expression for future thinking, foresight, planning, and creativity. It is well understood that memory alleviates forgetting, and that memorizing all examples is not only not scalable, but also not biologically principled.

It is understood that human memory does not offer perfect recall with full details. We can also conjecture it is biologically intractable to reactivate all memory cells for recall due to a finite synaptic-bandwidths. We also note that all animals, particularly all mammals, experience sleep, which consists of alternating sleep stages with characteristic electrophysiological features, across the sleep period. Sleep plays a critical role in supporting learning and relearning through memory consolidation [10].

From such biological principles on memory transfer, we derive an AI system with generative memory mechanism. Using replay and RL, we support learning from salient memories for lifelong learning. Using a generative memory, we address the ability to scale across lifelong events. We formulate an encoding method for the generative memory to separate the latent space such that learning can occur: (1) when the tasks or behavior is not well-defined, (2) when the number of tasks is unknown.

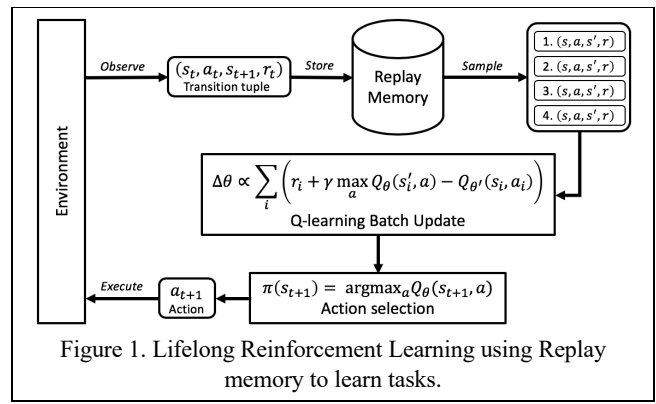

Figure 1. Lifelong Reinforcement Learning using Replay memory to learn tasks.

[†] Alternative email address: sek.chai@gmail.com.

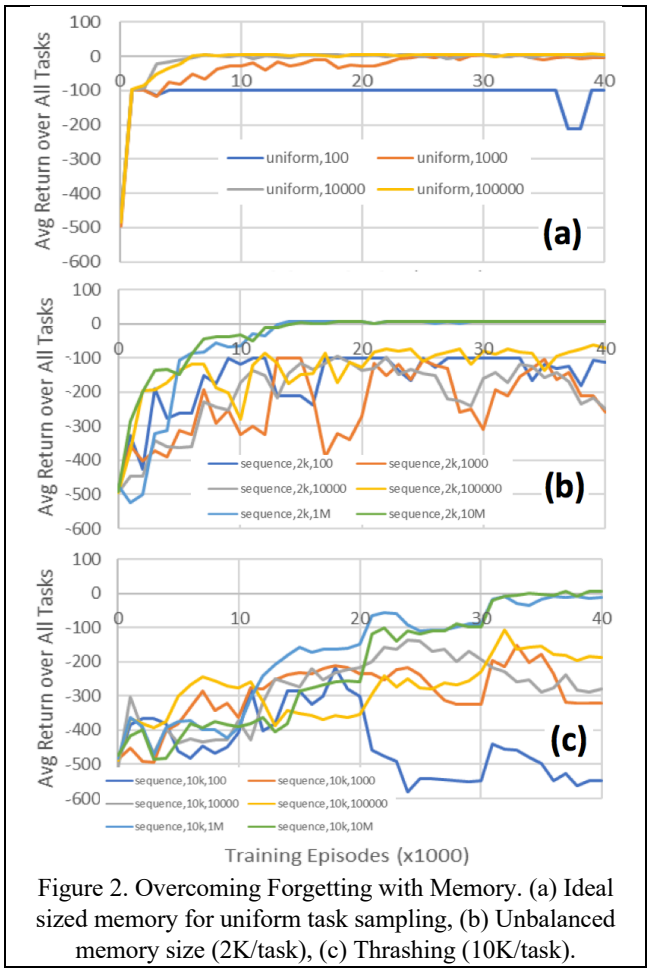

Figure 2. Overcoming Forgetting with Memory. (a) Ideal sized memory for uniform task sampling, (b) Unbalanced memory size (2K/task), (c) Thrashing (10K/task).

## 2 Generative Memory and Scalability

We first provide simulation results (Figure 2) to show the relationship between memory size and catastrophic forgetting, which motivates the need to have a scalable approach with sufficient memory sizes. For small number of tasks and short durations, a small memory works with simple uniform task sampling. However, larger sized memories are needed when tasks and duration between tasks increases [11]. Memory trashing examples (Figure 2c) are indicative of forgetting (even for a small number of tasks).

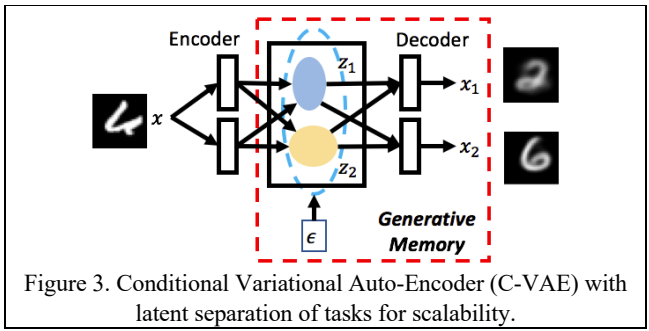

Figure 3. Conditional Variational Auto-Encoder (C-VAE) with latent separation of tasks for scalability.

In our approach, a generative memory enables sampling from all task with joint distribution over all tasks and instances. As shown in Figure 3, we use a conditional variational auto-encoder (C-VAE) [12], with generative properties to allows for a more scalable memory subsystem beyond a simple FIFO or cache using in today's RL approaches. To directly address scalability, we have developed an approach where the memory recall is agnostic to task labels. We use an embedding that forces concept separation (e.g. quantifying if memories are similar or different so that they can be recalled appropriately in the future). The equation governing the separation consists of a mathematical term (cosine function, in red) as part of the overall lost function:

$$\min_{\theta,\phi} -D_{KL}[q_\phi(z|x)||p_\theta(z)] + E_{q_\phi(z|x)}[w.\log p_\theta(x|z)] + \lambda \sum_{i,j} |\cos(\angle \mu_i, \mu_j)|$$

## 3 Experiments and Results

Our C-VAE is a generative memory with ability to learn concepts that are different, and to reinforce those that are similar. Figure 4 shows the ability to separate the latent space (e.g. $z_1$ and $z_2$ in Figure 3) for each of the encoders in the C-VAE. Using the $\lambda$ and $\theta$ parameters (equation above) to guide the encoding, we arrive at a generative memory system for lifelong RL, whereby the task embedding is accomplished in an unsupervised manner. Our approach supports a sub-linear growth in learning parameters and can support smooth transition across the latent space during adaptation to surprises.

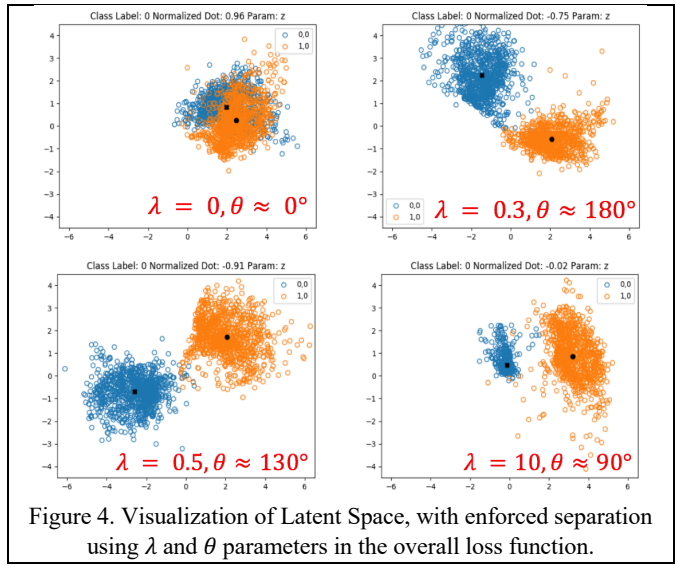

Figure 4. Visualization of Latent Space, with enforced separation using $\lambda$ and $\theta$ parameters in the overall loss function.

Using the MNIST dataset, we can show that the C-VAE and latent separation in the encoding. For example, as shown in Figure 3, we can show results where different output digits are encoded based on training on the same input datasets. We can show various other results whereby different lifelong RL behavior can be observed.

## 3    Conclusion

Memory is a key element of cognitive function and it is central to continual learning system. Our approach uses a generative memory in reinforcement learning framework to generate batch replay samples for training. Such an approach offers robust lifelong learning mechanisms to address issues with catastrophic forgetting. Our effort is focused specifically on memory transfer with basis on current understanding of biologically multi-stage memory consolidation.

## 4    Acknowledgements

This material is based upon work supported by the Lifelong Learning Machines (L2M) program of the Defense Advanced Research Projects Agency (DARPA) under Contract No. HR0011-18-C-0051. Any opinions, findings and conclusions or recommendations expressed in this material are those of the author(s) and do not necessarily reflect the views of DARPA. Special thanks to Dr. Hava Siegelmann, DARPA L2M Program Manager, for her support and guidance in the program.